# Automatic Question Generation for Intuitive Learning Utilizing Causal Graph Guided Chain of Thought Reasoning

Nichoas X. Wang

Stellar Learning Technologies
San Jose, USA
nicholas@stellarlearning.app

Neel V. Parpia

Stellar Learning Technologies
San Jose, USA
neel@stellarlearning.app

Aaryan D. Parikh

Stellar Learning Technologies
San Jose, USA
aaryan@stellarlearning.app

Aggelos K. Katsaggelos

Northwestern University
Evanston, USA
a-katsaggelos@northwestern.edu

*Abstract*— Intuitive learning plays a vital role in building deep conceptual understanding, particularly in STEM education, where students often grapple with abstract and interdependent ideas. Automatic question generation has emerged as an effective strategy to support personalized and adaptive learning. However, its effectiveness is limited by hallucinations in large language models (LLMs), which can produce factually incorrect, ambiguous, or pedagogically inconsistent questions. To address this challenge, we propose a novel framework that combines causal-graph-guided Chain-of-Thought (CoT) reasoning with a multi-agent LLM architecture to ensure the generation of accurate, meaningful, and curriculum-aligned questions. In this approach, causal graphs offer an explicit representation of domain knowledge, while CoT reasoning enables structured, step-by-step traversal through related concepts. Dedicated LLM agents handle specific tasks such as graph pathfinding, reasoning, validation, and output, all operating under domain constraints. A dual validation mechanism—at both the conceptual and output stages—substantially reduces hallucinations. Experimental results show up to a 70% improvement in quality over reference methods and yielded highly favorable outcomes in subjective evaluations.

*Keywords—Automatic Question Generation, Intuitive Learning, Causal Graph, Chain of Thought, LLM.*

## I. Introduction

Intuitive learning plays a critical role in bridging the gap between knowledge delivery and deep understanding. Unlike traditional approaches that emphasize rote memorization, intuitive learning aligns instruction with natural cognitive processes, encouraging exploration, contextualization, and step-by-step reasoning. This approach is particularly vital in STEM education, where abstract theories and complex problem-solving often hinder student progress. By fostering conceptual clarity over surface-level recall, intuitive learning empowers students to build lasting, transferable knowledge. As educational technologies evolve, integrating intuitive learning into AI-powered systems offers a promising path toward more personalized, engaging, and effective instruction. Recent studies [1–10] show how Large Language Models (LLMs) and generative AI can support this shift by creating dynamic, visual, and interactive learning experiences that help learners grasp complex ideas with greater ease and confidence.

Automatic question generation [11–15] is a cornerstone of intuitive learning, offering real-time, adaptive support that reinforces key concepts and challenges learners at the appropriate level. Unlike static assessments, AI-driven systems can tailor questions with scaffolded hints and varying difficulty, promoting deeper reasoning and independent thinking. However, LLM hallucinations [16] pose significant challenges, such as generating factually incorrect, ambiguous, or curriculum-inappropriate questions. These issues can mislead learners, reinforce misconceptions, and undermine trust in the educational tool, compromising both comprehension and assessment validity. Ensuring reliability and alignment with learning objectives requires structured approaches to mitigate such risks.

To address these challenges, we propose a novel method that combines causal-graph-guided Chain-of-Thought (CoT) reasoning with a multi-agent LLM framework for automatic question generation. In this approach, causal graphs provide a structured map of concept dependencies within a domain, while CoT reasoning guides logical progression through these concepts. Each LLM agent in the system is assigned a specialized role—such as pathfinding, reasoning, validation, or output—operating under domain-specific constraints to maintain accuracy. Dual validation steps ensure both the coherence of conceptual paths and the correctness of generated questions. By grounding generation in structured knowledge and reasoning, this approach significantly reduces hallucinations and enhances the pedagogical soundness, alignment, and reliability of AI-generated questions.

## II. Causal Graph for Intuitive Learning

Causal graphs—also known as causal knowledge maps or concept dependency graphs—hold strong potential for enabling intuitive learning. While they have already been leveraged in graph-based knowledge representation [1], their application in automatic question generation offers equally compelling benefits. Their key advantages include:

- Explicit representation of dependencies: Causal graphs illustrate how concepts are logically or causally connected (e.g., "Newton's Second Law → motion analysis → energy conservation"), making knowledge structures more transparent.

- Support for adaptive sequencing: They enable personalized learning by organizing content around prerequisite relationships, allowing for dynamic curriculum design and progression.

- Facilitation of question generation: Traversing the graph—either forward or backward—can generate context-aware, scaffolded questions that reflect a learner's current understanding.

- Targeted error diagnosis: Learner misconceptions can be traced to specific nodes or links, allowing for focused interventions and tailored remediation.

As shown in Fig. 1, causal graphs offer a powerful way to represent conceptual relationships across diverse academic domains. In physics, for example, concepts such as force, acceleration, velocity, and energy form clear causal links. In microeconomics, factors like consumer preferences, income, and supply drive changes in demand, which then influence equilibrium price and economic surplus. Many disciplines—especially in physics, mathematics, and engineering—can benefit from using causal graphs to mirror textbook derivations and problem-solving sequences. In fields like economics and the social sciences, causal models help structure interdependent variables. In biology and medicine, they effectively represent pathways such as "symptom → diagnosis → treatment" or "gene → protein → phenotype." Even in literature, philosophy, and history, causal graphs can loosely capture temporal or thematic sequences (e.g., "event A leads to movement B").

Causal graphs also enable structured and contextual question generation. Different traversal strategies yield distinct types of questions:

- Forward traversal: *"What follows from this principle?"*

- Backward traversal: *"What must be true for this result to occur?"*

- Branching: *"Which conditions affect this outcome?"*

- Misconception injection: *"What happens if this node is misunderstood?"*

For instance, in the Mechanics graph (as shown in Fig. 1), forward traversal from *Force → Acceleration → Velocity* might generate the question, *"If a constant force is applied to an object, how does its velocity change over time?"* A backward traversal from *Work ← Force* might yield, *"What factors determine the amount of work done on an object?"* Conditional reasoning could prompt, *"How would increasing mass affect the acceleration of an object under a constant force?"* And a full causal chain might support a question like, *"Describe how force leads to changes in kinetic energy in an object."*

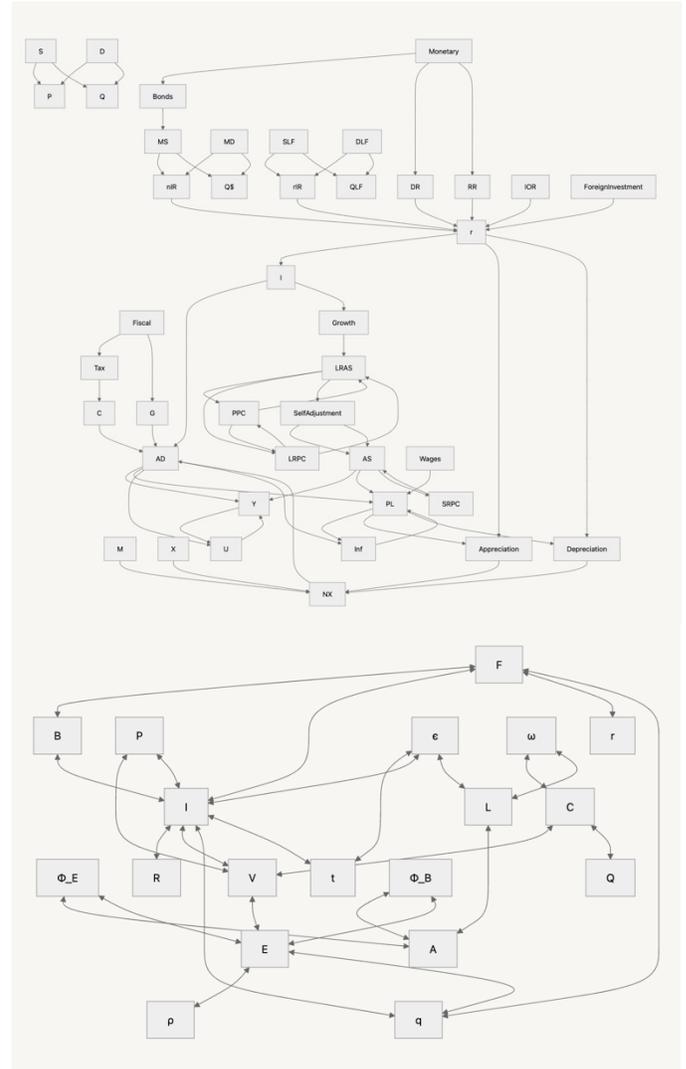

Figure 1. Sample Causal Graph in Macroeconomics and E&M

## III. CAUSAL GRAPH GUIDED CHAIN-OF-THOUGHT REASONING FOR QUESTION GENERATION

Integrating causal graphs with chain-of-thought (CoT) reasoning creates a powerful foundation for generating deeper, more coherent, and instructionally effective questions in education. Causal graphs define a domain-specific structure, mapping how concepts are logically or causally connected. This structured layout serves as a roadmap for navigating knowledge in a purposeful sequence. Chain-of-thought reasoning, meanwhile, mirrors the human approach to solving problems step by step. When combined, each node or link in a causal graph can correspond to a step in the reasoning process, enabling the generation of questions that reflect both conceptual structure and cognitive flow.

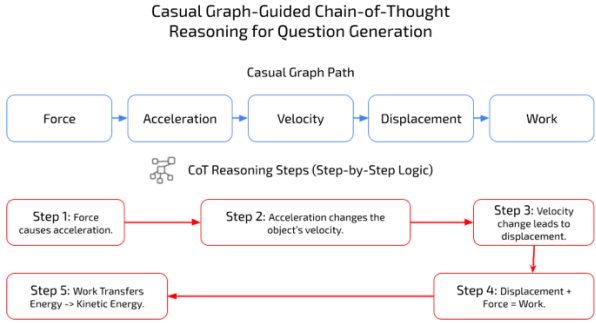

Figure 2. An example of relationship between Causal graph and Chain of Thought (CoT) in question generation

As illustrated in Fig. 2, causal graphs define the key concepts within a subject and how they are logically or causally connected in a specific sequence. In parallel, chain-of-thought (CoT) reasoning provides a step-by-step explanation of how to move through these connections to solve a problem or understand a concept. When combined, the causal graph provides the structure—*what to reason about and in what order*—while the CoT provides the *how*—the logical reasoning path that connects each concept.

In this work, we propose that this integrated approach—Causal Graph–guided Chain-of-Thought reasoning—can be effectively used for automatic question generation. By aligning the traversal of a causal graph with reasoning steps, a system can mimic the structure and logic found in existing educational materials, such as AP exam questions or textbook exercises. For instance, a simple path in the graph might generate a basic question, while expanding the graph to include more interconnected concepts can lead to more complex, multi-step questions. In this way, questions can be generated at varying levels of difficulty by adjusting the size and depth of the subgraph used, enabling a scalable, curriculum-aligned approach to intelligent tutoring and assessment.

As shown in Fig. 3, the causal graph–guided chain-of-thought (CoT) reasoning framework is implemented through a coordinated system of Large Language Model (LLM) agents, each responsible for a specific task:

- Causal Graph Pathfinder Agent: Converts an input question into a corresponding path within the causal graph.
- Causal Graph Path Expansion Agent: Derives new conceptual paths in the causal graph based on existing question paths from training data.
- Causal Graph Path Validation Agent: Evaluates whether a proposed path is conceptually and logically valid.
- CoT-Based Question Generation Agent: Generates new questions using training data, CoT reasoning steps, and the expanded causal graph path.
- Question Validation Agent: Assesses the quality and correctness of the generated question.
- Question Output Agent: Delivers the validated question to the end user.

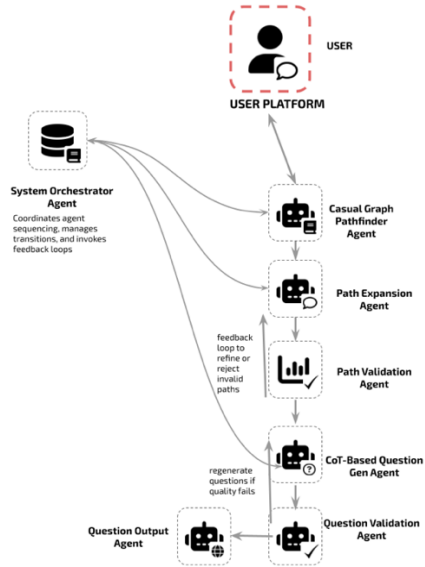

Figure 3. The implementation of causal graph–guided chain-of-thought framework with a coordinated system of LLM agents

This multi-agent design ensures that reasoning follows a structured and validated conceptual flow. By incorporating dual validation steps—one for the causal graph path and another for the generated question—the system significantly reduces the likelihood of LLM hallucination, resulting in more reliable and pedagogically sound outputs.

IV. EXPERIMENTAL RESULTS

The proposed framework has been integrated into a public online learning platform called *Stellar* [5], which currently offers over 40 courses and has been used by more than 5,000 students. To evaluate its performance, we compared *Stellar* with two reference systems: *ChatGPT* [6] and the commercial online learning product *Knowt* [17]. The evaluation was conducted using the following objective metrics:

- Flesch-Kincaid Grade Level: A standard readability metric that estimates the U.S. school grade level required to understand the question, based on sentence length and word complexity.
- Key Points: The number of domain-specific terms and learning objectives addressed by each question, reflecting its conceptual depth.
- Solution Quality: An indicator of question complexity, measured by the number of logical steps required in a complete solution.

As shown in Fig. 4, Stellar consistently outperforms both ChatGPT and Knowt across all evaluated metrics (normalized for clarity), achieving the highest overall quality scores. In particular, Stellar demonstrates a clear advantage in the Key Points category, thanks to its causal graph–based approach, which enables more comprehensive connections among knowledge points—leading to deeper conceptual understanding. This strength also translates to superior solution quality, as the reasoning process is guided by causal graph–driven Chain-of-Thought (CoT), allowing for greater logical control. The overall score is computed using weighted metrics—for example, (0.3, 0.3, 0.4)—and, as illustrated, Stellar's score exceeds Knowt's by 70%.

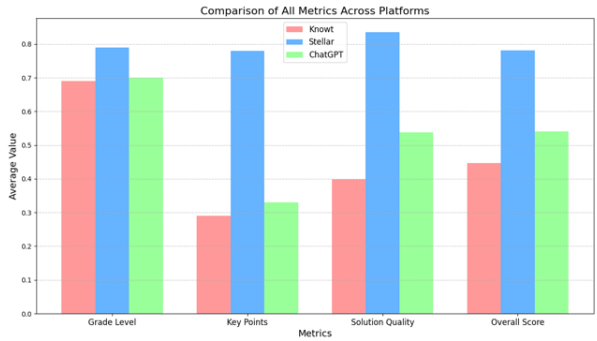

Figure 4. Objective test outcome of comparing Stellar with Knowt and ChatGPT

Fig. 5 provides a more detailed comparison of overall scores, including their relationship to the number of questions and the distribution of those scores. All systems exhibited reasonably stable performance across the full set of questions. Notably, Stellar consistently outperformed the two reference methods, with performance gains reaching up to 70%.

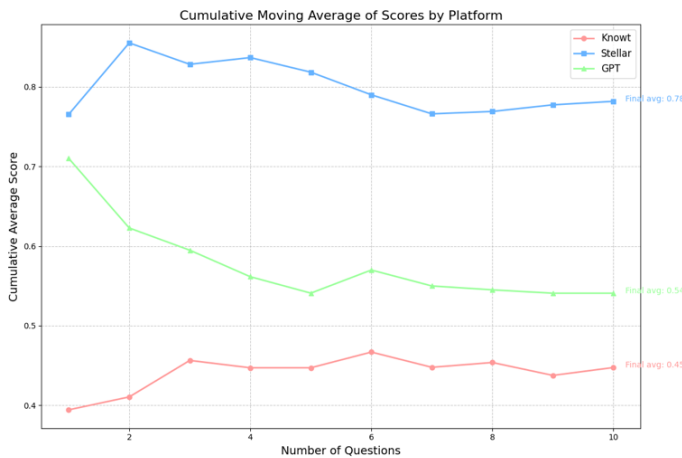

(a) The cumulative score across multiple questions

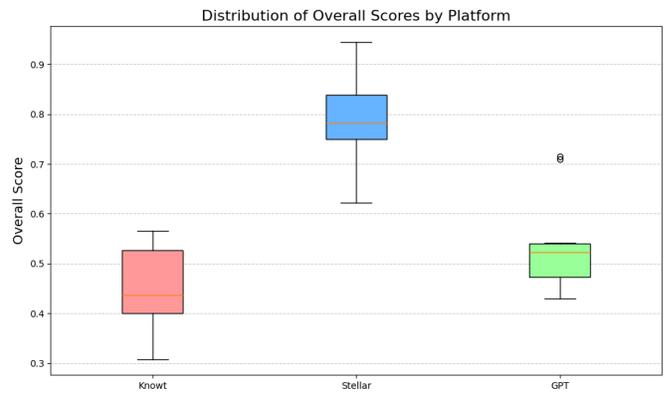

(b) The distribution of overall scores

Figure 5. Overall score comparison

Fig. 6 presents a set of representative sample questions generated by Stellar alongside those produced by the reference systems. A clear difference can be observed in the depth, clarity, and comprehensiveness of the questions produced by Stellar. This distinction highlights the effectiveness of the causal graph-guided chain-of-thought (CoT) reasoning employed by Stellar during the question generation process. By leveraging structured knowledge pathways, Stellar can create questions that better capture underlying concepts, integrate multiple knowledge points, and promote a more intuitive and connected understanding—surpassing the more surface-level questions generated by baseline methods.

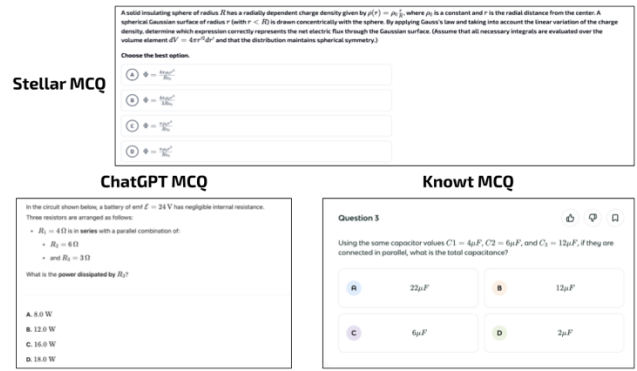

Figure 6. Sample questions generated by Stellar and reference solutions

In order to assess the quality of the generated questions based on user perception, we performed a user study with 25 subjects who used the system over a range of STEM and humanities subjects. The response was overwhelmingly positive: 17 users voted with 5 stars, 5 voted with 4 stars, 2 voted with 3 stars, and 1 voted with 2 stars as shown in Fig. 7. Users repeatedly emphasized the depth, clarity, and contextual significance of the questions. Numerous users commented that the step-by-step reasoning built into the prompts seemed "natural" and "how they reason through things." A number complimented how the questions scaffolded ideas in a progressive way, making difficult material more accessible. The lower ratings were mostly attributed to occasional question difficulty mismatches or redundancy in wording. Nevertheless, more than 90% of participants indicated the generated questions were far more useful than those from other products. In conclusion, the subjective feedback adds validity that causal-graph-guided Chain-of-Thought reasoning increases the instructional effectiveness and perceived worth of AI-generated questions.

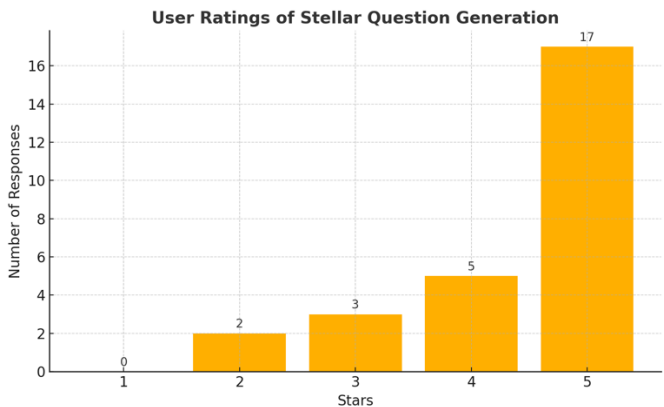

Figure 7. Distribution of Subjective Test Ratings

V. CONLCUSION

In this work, we introduce a novel framework that integrates causal-graph-guided Chain-of-Thought (CoT) reasoning with a multi-agent LLM architecture to enable reliable and pedagogically sound automatic question generation for intuitive learning. This framework has been deployed within *Stellar* [5], an intuitive learning platform launched in November 2024. Since its release, Stellar has supported over 5,000 students across more than 40 courses, with approximately 20% of users engaging for over an hour on a near-daily basis. The proposed approach addresses a long-standing challenge in educational AI—the hallucination problem in LLMs—by using causal graphs to explicitly represent domain knowledge and CoT reasoning to guide logical, step-by-step question generation aligned with concept dependencies. While the framework has shown strong results in STEM and other structurally defined domains, future research is needed to explore its adaptability to disciplines like English literature, where knowledge is less easily modeled through causal relationships.

ACKNOWLEDGMENT

We extend our heartfelt thanks to all colleagues and volunteers whose invaluable contributions were essential to the success of Stellar and this paper.